%% file: acl_latex.tex
\pdfoutput=1
\documentclass[11pt]{article}

\usepackage[final]{acl}

\usepackage{times}
\usepackage{latexsym}

\usepackage[T1]{fontenc}
\usepackage[utf8]{inputenc}

\usepackage{microtype}

\usepackage{inconsolata}

\usepackage{graphicx}

\usepackage{cite}
\usepackage{amsmath,amssymb,amsfonts}
\usepackage{algorithmic}
\usepackage{graphicx}
\usepackage{textcomp}
\usepackage{multicol}
\usepackage{multirow}
\usepackage{xcolor}
\usepackage{pifont}

\usepackage{booktabs, multirow} % for borders and merged ranges
\usepackage{soul}% for underlines
\usepackage{changepage,threeparttable} % for wide tables
\usepackage{amsmath,amssymb,amsfonts}
\usepackage{textcomp}
\usepackage{xcolor}
\usepackage{amsmath,tabularx}
\usepackage{makecell}
\usepackage{tablefootnote}
\usepackage{changepage,threeparttable}
\usepackage{hyperref}
\usepackage{times}

\title{Speech Discrete Tokens or Continuous Features? A Comparative Analysis for Spoken Language Understanding in SpeechLLMs}

\author{%
  Dingdong Wang,  Junan Li, Mingyu Cui, Dongchao Yang, Xueyuan Chen, Helen Meng
  \\
  The Chinese University of Hong Kong \\
  \texttt{dingdongwang@link.cuhk.edu.hk} 
}

\begin{document}
\maketitle
\begin{abstract}
With the rise of Speech Large Language Models (SpeechLLMs), two dominant approaches have emerged for speech processing: discrete tokens and continuous features. Each approach has demonstrated strong capabilities in audio-related processing tasks. However, the performance gap between these two paradigms has not been thoroughly explored. To address this gap, we present a fair comparison of self-supervised learning (SSL)-based discrete and continuous features under the same experimental settings. We evaluate their performance across six spoken language understanding-related tasks using both small and large-scale LLMs (Qwen1.5-0.5B and Llama3.1-8B). We further conduct in-depth analyses, including efficient comparison, SSL layer analysis, LLM layer analysis, and robustness comparison. Our findings reveal that continuous features generally outperform discrete tokens in various tasks. Each speech processing method exhibits distinct characteristics and patterns in how it learns and processes speech information. We hope our results will provide valuable insights to advance spoken language understanding in SpeechLLMs.
\end{abstract}

\input{sections/1-introduction}

\input{sections/3-pipeline}

\input{sections/4-experiment}

\input{sections/5-discussion}

\input{sections/6-conclusion}

\section{Limitations}

Due to limited computational resources, we were unable to incorporate a broader range of LLM backbones in our study. Expanding the number of models would have provided a more extensive comparison, potentially uncovering additional insights into the strengths and weaknesses of discrete tokens and continuous features. Furthermore, due to the extensive number of tasks and domains covered in our evaluation, we focused primarily on the LibriSpeech 960-hour dataset for our ablation study. While this allows for fair comparisons between models on a consistent dataset, it does not take into account the large-scale fine-tuning typically employed in state-of-the-art (SOTA) settings for each specific downstream task. This omission may limit the applicability of our findings to real-world settings, where task-specific fine-tuning is often crucial for achieving optimal performance.

\bibliography{custom}

\newpage

\appendix

\input{sections/7-appendix}

\end{document}

%% file: sections/1-introduction.tex
\section{Introduction}

Learning speech representations that are robust and effective is a key challenge in modern speech processing systems. Recent advances in Speech Large Language Models (SpeechLLMs), especially systems such as GPT-4o, have captured significant attention in the speech domain~\citep{ji2024wavchat,arora2025landscape,cui2025recentadvancesspeechlanguage,peng2025surveyspeechlargelanguage,Yin_2024,caffagni2024revolutionmultimodal}. With the advancement of SpeechLLMs, two main types of speech representations are employed as inputs: continuous features and discrete speech tokens.

Using speech continuous features is an intuitive approach~\citep{chu2023qwen,chen2023lauragpt,gong2023joint,wang2023blsp,tang2023salmonn,gong2023listen,wang2025inserter} for integrating speech signals with large language models (LLMs). This type of representation is typically obtained from certain layers of a self-supervised learning (SSL) model or a speech encoder module. Specifically, in SLAM-ASR~\citep{ma2024embarrassingly}, continuous features are obtained from the final layer of the HuBERT model. In models like the Qwen-Audio series~\citep{chu2023qwen,Qwen2-Audio} and SALMONN~\citep{tang2023salmonn}, raw speech is transformed into high-dimensional embeddings using the Whisper encoder~\citep{radford2023robust} and adapted for LLMs through an adapter module. These continuous features preserve the audio signal’s richness and have shown strong performance in speech understanding~\citep{ji2024wavchat}. However, since autoregressive LLMs like GPT~\citep{gpt4} and LLaMA~\citep{dubey2024llama} are naturally designed to work with discrete text tokens, this has inspired researchers to explore representing speech as sequences of discrete tokens.

Recent interest in speech discrete tokens has grown rapidly, driven by their compatibility with large language models that operate on text tokens. Unlike continuous features, which require a speech adapter for modality alignment, discrete tokens can be directly integrated into the LLM's vocabulary. Several works have used discrete tokens in SpeechLLMs~\citep{zhang2023speechgpt, rubenstein2023audiopalm, wang2023viola, wang2024speechx, mitsui2024pslmparallelgenerationtext, nguyen2024spiritlminterleavedspoken, défossez2024moshispeechtextfoundationmodel}. For instance, both AudioPaLM~\citep{rubenstein2023audiopalm} and SpeechGPT~\citep{zhang2023speechgpt} use K-means clustering to discretize speech embeddings. More recent works, such as Mini-Omni series~\citep{xie2024miniomnilanguagemodelshear,xie2024miniomni2} and GLM-4-Voice~\citep{zeng2024glm4}, they explore tokenizer modules based on compression methods, leveraging encoder-decoder architectures with residual vector quantization (RVQ)~\citep{zeghidour2021soundstream} to extract discrete tokens.

Continuous-feature-based and discrete-token-based SpeechLLMs have each demonstrated promising performance across various speech-related tasks~\citep{ji2024wavchat,arora2025landscape}. However, despite their success, there has been no systematic and fair comparison between these two types of representation within the spoken language understanding (SLU) area. Intuitively, discrete tokens compress information through quantization, sacrificing precision for compactness, while continuous representations retain fine-grained acoustic and temporal details, which may affect the performance of discrete tokens when compared to continuous features. However, the extent and nature of the potential performance differences remain unclear. Moreover, in multimodal settings, it is still largely unknown whether LLMs interpret these two forms of speech input differently, and how such differences affect various downstream SLU-related task outcomes. Clarifying these distinctions is crucial for understanding each paradigm's limitations and guiding future advancements in architecture, tokenizer design, and training methodologies.

% Continuous-feature-based Speech LLMs and discrete-token-based Speech LLMs have each demonstrated promising performance in handling speech-related tasks. However, these two approaches have not yet been directly and thoroughly compared. In multimodal settings, whether there exists a difference in how LLMs understand these two distinct types of speech features, and the specific nature of these differences, remains rarely explored. Understanding this distinction is crucial for the advancement of the Speech LLM field. To address this gap, our research aims to conduct a fair and comprehensive comparison between continuous features and discrete tokens under consistent settings. 

To address this gap, we present a comprehensive benchmark comparing continuous features and discrete tokens under consistent experimental conditions. Our experiments utilize self-supervised learning (SSL) models for speech feature extraction, leveraging K-means clustering for discretization and using original embeddings as continuous features. This approach provides a straightforward way to compare these two paradigms. Specifically, we select HuBERT-Large~\citep{hsu2021hubert} and WavLM-Large~\citep{chen2022wavlm} as SSL models, and Qwen1.5-0.5B~\citep{bai2023qwen} and Llama3.1-8B~\citep{dubey2024llama} as LLM decoders. We aim to investigate the performance differences between discrete tokens and continuous features in spoken language understanding, evaluating both representations across a range of tasks, including automatic speech recognition, phoneme recognition, speech translation intent classification, keyword spotting, and emotion recognition.

Beyond benchmark comparison, we further conduct detailed analysis from multiple perspectives, including efficiency, SSL and LLM layer behaviour, and robustness. Across a range of tasks, our results show that continuous features generally outperform discrete tokens, with the magnitude of this gap varying by task. Interestingly, discrete tokens perform better on phonetic-level tasks, indicating their strength in capturing subword-level structure. Our SSL and LLM layer analyses also reveal distinct learning patterns for each representation across the model hierarchy. Each representation paradigm also demonstrates unique strengths: discrete tokens excel in data compactness and training efficiency, while continuous features offer superior robustness in diverse conditions. Our key contributions are as follows:

\begin{itemize}
    \item 
    % A comparative analysis of continuous features and discrete tokens is conducted across five semantic tasks (speech recoginition, phoneme recognition, speech translation, intent classification, and keyword spotting). 
    We present the first comprehensive benchmark comparing speech continuous features and discrete tokens for spoken language understanding (SLU) under consistent experimental conditions.
    % \item We assess other attributes (i.e. the training efficiency and data size scalability) of both continuous features and discrete speech tokens.
    \item We conduct an in-depth analysis of both representations, examining efficiency, robustness, and learning dynamics across SSL and LLM layers.
    % \item We investigate the potential factors behind the performance gap between continuous features and discrete tokens. 
    \item We identify the complementary strengths of each representation, offering practical guidance for future SpeechLLMs development.
\end{itemize}

%% file: sections/3-pipeline.tex
\section{Pipeline Design}

\begin{figure}[!t]
    \centering
    \includegraphics[width=0.45\textwidth]{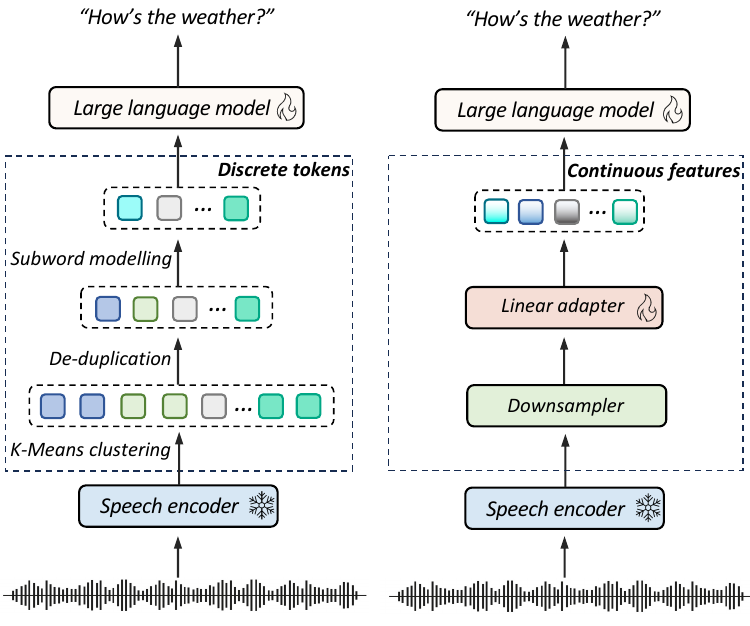}
    \caption{\small Architectures of two approaches for integrating speech into Large Language Models (LLMs). (Left) discrete token-based encoding. (Right) continuous feature processing.}
    \label{fig:arch}
\end{figure}

To systematically compare the performance of continuous and discrete tokens in SLU related tasks, we adopt two widely used~\citep{Qwen2-Audio,xu2025qwen25omnitechnicalreport,gong2023joint,wang2023blsp,zhang2023speechgpt,shon2024discreteslu,dekel2024exploring} speech processing pipelines, as illustrated in Fig.~\ref{fig:arch}: (a) a discrete token pipeline and (b) a continuous features pipeline.

\subsection{Speech Discretization}
In our approach, we utilize SSL models (HuBERT-Large~\citep{hsu2021hubert}, WavLM-Large~\citep{chen2022wavlm}) to generate discrete tokens. As shown in Fig.~\ref{fig:arch}(a), we first generate a sequence of high-dimensional feature vectors \( H = \{h_1, h_2, \dots, h_T\} \) from the audio waveform using an SSL model. Second, we apply \textbf{K-Means Clustering} to obtain discrete tokens \( Z = \{z_1, z_2, \dots, z_T\} \). These tokens now exist as discrete symbols that can be processed like text symbols in NLP, making it possible to use traditional NLP techniques.

After clustering, the discrete token sequence may contain redundant consecutive information. Thus we apply \textbf{De-duplication} to merge consecutive identical tokens to reduce redundancy, and finally use \textbf{Byte-Pair Encoding (BPE)} to enhance input tokens by combining frequent subsequences into shorter meta tokens \( Z' = \{z'_1, z'_2, \dots, z'_{T'}\} \), where \( T' < T \). Mathematically, the length reduction ratio \( \frac{T'}{T} \) typically ranges from 30\% to 60\%, depending on the K-means clustering granularity and the BPE vocabulary size.

\subsection{Processing Continuous Features}

For continuous features, we utilize the same speech encoders. As shown in Fig.~\ref{fig:arch}(b), the processing pipeline involves two steps: (1) \textbf{Downsampling}, where the feature sequence \( H = \{h_1, h_2, \dots, h_T\} \) is reduced by concatenating every \( k \) consecutive frame, producing \( Z^S = \{z_1^S, \dots, z_N^S\} \) with \( N = T/k \). This reduces computational complexity while preserving key information. (2) \textbf{Linear Adapter}: After downsampling, we use a single linear adapter
layer to map the embeddings to the LLM’s hidden size.

\subsection{Instruction-Tuning with LLMs}

We select six representative tasks related to spoken language understanding: Automatic Speech Recognition (ASR), Phoneme Recognition (PR), Keyword Spotting (KS), Emotion Recognition (ER), Spoken Intent Classification (IC), and Speech Translation (ST). For each task, we design specific prompts to guide the model in following instructions. Detailed prompt designs and task-related datasets information can be found in the appendix.

We perform instruction-tuning on the discrete and continuous inputs for these tasks using the Qwen1.5-0.5B~\citep{bai2023qwen} and Llama3.1-8B~\citep{dubey2024llama} models as decoders. Our goal is to explore how different downstream tasks and speech representations (discrete tokens vs. continuous features) interact with LLMs of varying scales, revealing insights into their performance in spoken language understanding.

%% file: sections/4-experiment.tex
\label{3-experiment}
\section{Experiments and Results}
\subsection{Experimental Setup}

To investigate the performance of discrete tokens compared to continuous features across various spoken language understanding related tasks, we conduct experiments using a range of datasets: LibriSpeech~\citep{panayotov2015librispeech}, GigaSpeech M-size~\citep{chen2021gigaspeech} and CHiME-4~\citep{vincent2016chime} for \textbf{ASR}, LibriSpeech 100-hour~\citep{panayotov2015librispeech} for \textbf{PR}, Speech Commands-v2~\citep{warden2018speech} for \textbf{KS}, SLURP~\citep{bastianelli2020slurp} for \textbf{IC}, and GigaST~\citep{ye2022gigast} for \textbf{ST}. To evaluate paralinguistic emotion information, we also include \textbf{ER} using the IEMOCAP dataset~\citep{busso2008iemocap}.  

For a fair comparison, we consistently extract features from the final layer of both HuBERT-Large~\citep{hsu2021hubert} and WavLM-Large~\citep{chen2022wavlm} at a 16,000 Hz sampling rate for both feature types. For LLM training on Qwen1.5-0.5B, we perform full-parameter fine-tuning for both discrete and continuous methods. For LLaMA3.1-8B, we employ LoRA~\citep{hu2021lora} for efficiency by injecting adapters (rank=8 and \( \alpha = 16 \)) into the projection layers for keys and queries in all self-attention layers. 

For the speech processing procedure of discrete token, we chose K-means with 2000 centroids and 6000 BPE vocabulary size for generating discrete tokens across all tasks. For continuous features, we apply a downsampling rate of \( \alpha = 2 \) and a single-layer linear adapter to project the embeddings to the LLM input space. All experiments were conducted under identical settings, with a learning rate of \( 1 \times 10^{-5} \), batch size of 32, using the AdamW optimizer with a weight decay of \( 10^{-2} \). The parameter settings for each method are optimized according to ablation studies in Sec.~\ref{subsec:ablation}.

\input{tables/main}

\subsection{Main Results}

In our experiments, we explore the performance of discrete tokens and continuous features in a range of tasks, comparing them across different LLM scales (Table~\ref{tab:main-qwen}). Overall, for both the Qwen1.5-0.5B~\citep{bai2023qwen} and LLaMA3.1-8B~\citep{dubey2024llama}, continuous features consistently outperform discrete tokens in most tasks. Notably, WavLM-Large's continuous features perform the best across Automatic Speech Recognition (ASR), Speech Translation (ST), and Emotion Recognition (ER) tasks on the LLaMA3.1-8B model, while HuBERT-Large's continuous features show the best performance in Keyword Spotting (KS) and Intent Classification (IC) tasks on the same LLM.

The performance gap between discrete tokens and continuous features tends to increase as the LLM model size grows. For example, in Speech Translation, the BLUE score of continuous features on LLaMA3.1-8B is significantly higher than that of discrete tokens, showing the increasing advantage of continuous features with larger models. In certain tasks and datasets, such as CHiME-4 (a noisy background ASR dataset), discrete tokens show a decline in ASR performance, whereas continuous features maintain more stable performance. Additionally, in Emotion Recognition, as the LLM decoder size increases, discrete token performance remains consistently poor, with accuracy significantly lower than that of continuous features. For example, WavLM-Large's discrete tokens achieve an accuracy of 37.98\% on Qwen1.5-0.5B and 36.12\% on LLaMA3.1-8B, while the corresponding continuous features achieve 59.45\% on Qwen1.5-0.5B and 65.87\% on LLaMA3.1-8B model.

Interestingly, in the Phoneme Recognition (PR) task, discrete tokens outperform continuous features across all model scales. Specifically, WavLM-Large's discrete tokens achieve the best performance on LLaMA3.1-8B, with 7.02\% phoneme error rate (PER) score. This improvement likely stems from the discrete token representation, which aligns more closely with the phoneme-level structure of speech, facilitating easier learning than continuous features.

\begin{figure*}[!t]
    \centering
    \includegraphics[width=1\linewidth]{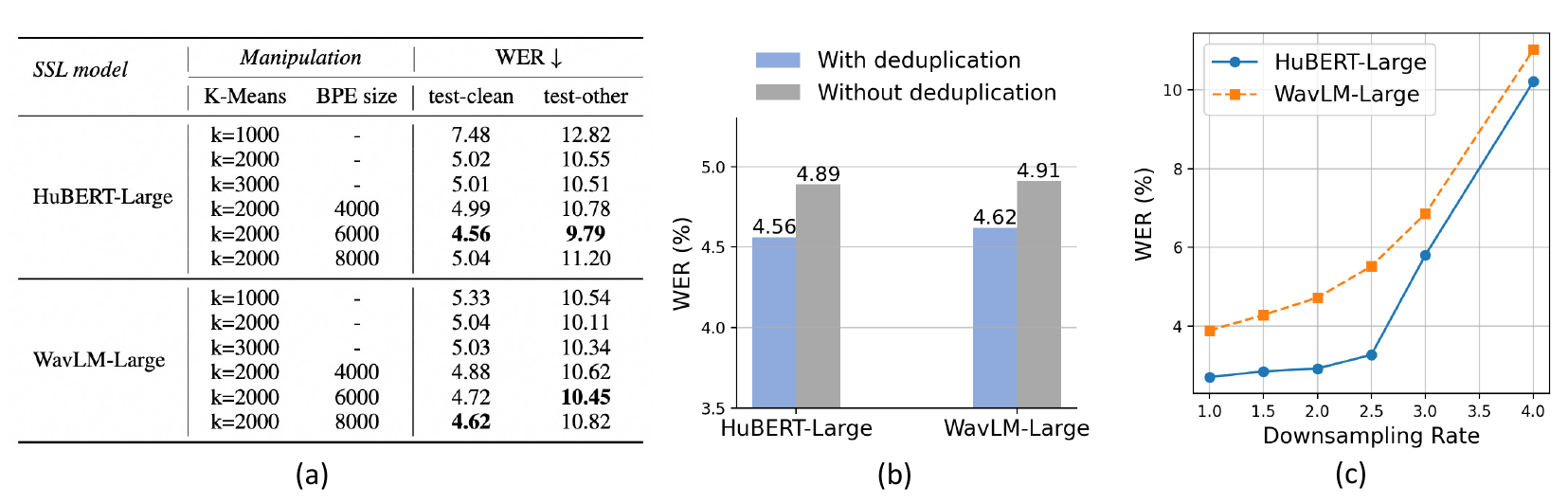}
    % \vspace{-4mm}
    \caption{\small Ablation studies: (a) Effect of K-means centroids and BPE size on WER; (b) Impact of deduplication method; (c) WER variation with different downsampling rates.}
    \label{fig:ablation}
    \vspace{-2mm}
\end{figure*}

\subsection{Ablation Study}
\label{subsec:ablation}

Our ablation study is conducted using the LibriSpeech 960-hour dataset with the Qwen1.5-0.5B model, and evaluations are performed according to the test-clean subset.

\paragraph{K-Meaning Clustering and BPE Size Settings.}
We investigate the effect of varying K-means clustering sizes and BPE (Byte Pair Encoding) vocabulary sizes on ASR performance. As shown in Fig.~\ref{fig:ablation}(a), increasing the number of centroids generally improves model performance. BPE-based subword modelling can further improve Word Error Rate (WER) results by reducing token sequence length while preserving key semantic information. Notably, the combination of \( k = 2000 \) centroids and 6000 BPE vocabulary size achieves a balanced trade-off between performance gains and computational efficiency. We fix this setting for all datasets to maintain discrete tokens consistency across tasks.

\paragraph{De-duplication Processing.}

As shown in Fig.~\ref{fig:ablation}(b), we examine the impact of de-duplication method applied during discretization. Our experiments show that using deduplication significantly improves WER performance, with WavLM-Large showing a reduction from 4.91\% to 4.62\%. The de-duplication process condenses consecutive identical tokens into a single token, reducing redundancy and improving information transmission efficiency.

\paragraph{Downsampling Settings.}

As shown in Fig.~\ref{fig:ablation}(c), we investigate the effect of downsampling rates on continuous features. As the rate increases, we observe a significant increase in WER for both HuBERT-Large and WavLM-Large. When the downsampling rate reaches 3, the WER starts to increase more sharply, suggesting a diminishing return in performance due to excessive downsampling. We select a downsampling rate of 2 as the optimal balance between efficiency and performance for all continuous feature settings.

%% file: tables/main.tex
\begin{table*}[t]
\centering

\footnotesize
\vspace{-0.2cm}
\renewcommand{\arraystretch}{1.5}  % 使表格行高变大
\setlength{\tabcolsep}{1.5pt}
\begin{tabular}{ll|c|c|c|c|c|c|c|c}
\toprule

\multirow{2}{*}{\textit{SSL model}} & \multirow{2}{*}{\textit{Token type}} & \multicolumn{3}{c}{ASR (WER$\downarrow$)} & \multicolumn{1}{|c|}{PR (PER$\downarrow$)} & \multicolumn{1}{c|}{ST (BLEU$\uparrow$)} & \multicolumn{1}{c|}{KS (ACC$\uparrow$)} & \multicolumn{1}{c|}{IC (ACC$\uparrow$)} & \multicolumn{1}{c}{ER (ACC$\uparrow$)} \\ \cmidrule{3-4} \cmidrule{5-10}

& & \makecell{LS \\ (clean$\vert$other)} & \makecell{GigaSpeech\\(test)} & \makecell{CHiME4\\(test)} & \makecell{LS-100\\(clean)} & \makecell{GigaST\\(En-Zh$\vert$En-De)} & \makecell{SC\\(test$\vert$val)} & \makecell{SLURP\\(test)} & \makecell{IEMOCAP\\(test)} \\ \midrule

\multicolumn{10}{c}{\textit{Qwen 1.5-0.5B}} \\
\midrule

HuBERT & \multirow{2}{*}{Discrete} & 4.56 / 9.79 & 19.40 & 13.35 & 9.69 & 22.75 / 20.14 & 93.70 / 93.85 & 57.04 & 38.65  \\
WavLM &  & 4.72 / 10.45 & 16.34 & 12.94 & 9.64 & 24.62 / 21.22 & 92.87 / 92.45 & 59.96 & 37.98 \\ \hline

HuBERT & \multirow{2}{*}{Continuous} & 4.91 / 6.43 & 17.45 & 8.62 & 12.84 & 26.63 / 25.42 & 95.38 / 95.70 & 76.84 & 56.72 \\
WavLM &  & 2.92 / 4.61 & 13.96 & 8.68 & 12.62 & 29.44 / 28.12 & 97.76 / 97.36 & 81.35 & 59.45 \\

\midrule
\multicolumn{10}{c}{\textit{Llama 3.1-8B}} \\
\midrule

HuBERT & \multirow{2}{*}{Discrete} & 2.56 / 6.49 & 12.86 & 10.56 & 7.85 & 26.64 / 25.13 & 96.75 / 96.69 & 63.44 & 39.84  \\
WavLM &  & 2.96 / 7.48 & 13.35 & 9.13 & \textbf{7.02} & 28.62 / 26.87 & 97.92 / 98.17 & 66.96 & 36.12 \\ \hline

HuBERT & \multirow{2}{*}{Continuous} & 1.76 / 4.58 & 9.04 & 5.72 & 9.83 & 32.02 / 34.42 & \textbf{99.74} / \textbf{98.59} & \textbf{86.84} & 64.54 \\
WavLM &  & \textbf{1.65} / \textbf{4.22} & \textbf{8.86} & \textbf{5.43} & 10.44 & \textbf{35.17} / \textbf{37.20} & 97.34 / 98.26 & 85.57 & \textbf{65.87} \\

\bottomrule
\end{tabular}
% \\ \scriptsize{feel free to edit here}
\caption{\small Comparison benchmark of discrete tokens and continuous features on various tasks. Discrete tokens use K-means (2000 clusters) with BPE size 6000 for all tasks. For LibriSpeech (LS) datasets, we evaluate test-clean (clean) and test-other (other) set. SC stands for Speech Commands-v2 dataset.}
\label{tab:main-qwen}
\end{table*}

%% file: sections/5-discussion.tex
\label{4-discussion}
\section{Discussion}

To provide a deeper understanding of each speech processing paradigm (continuous v.s. discrete), we conduct an in-depth analysis across several key areas, including efficiency comparisons, SSL layer contributions, LLM layer behaviours, and robustness in noisy conditions.

\subsection{Efficiency Comparison}

\paragraph{Data Efficiency}

\begin{figure*}[!t]
    \centering
    \includegraphics[width=1\textwidth]{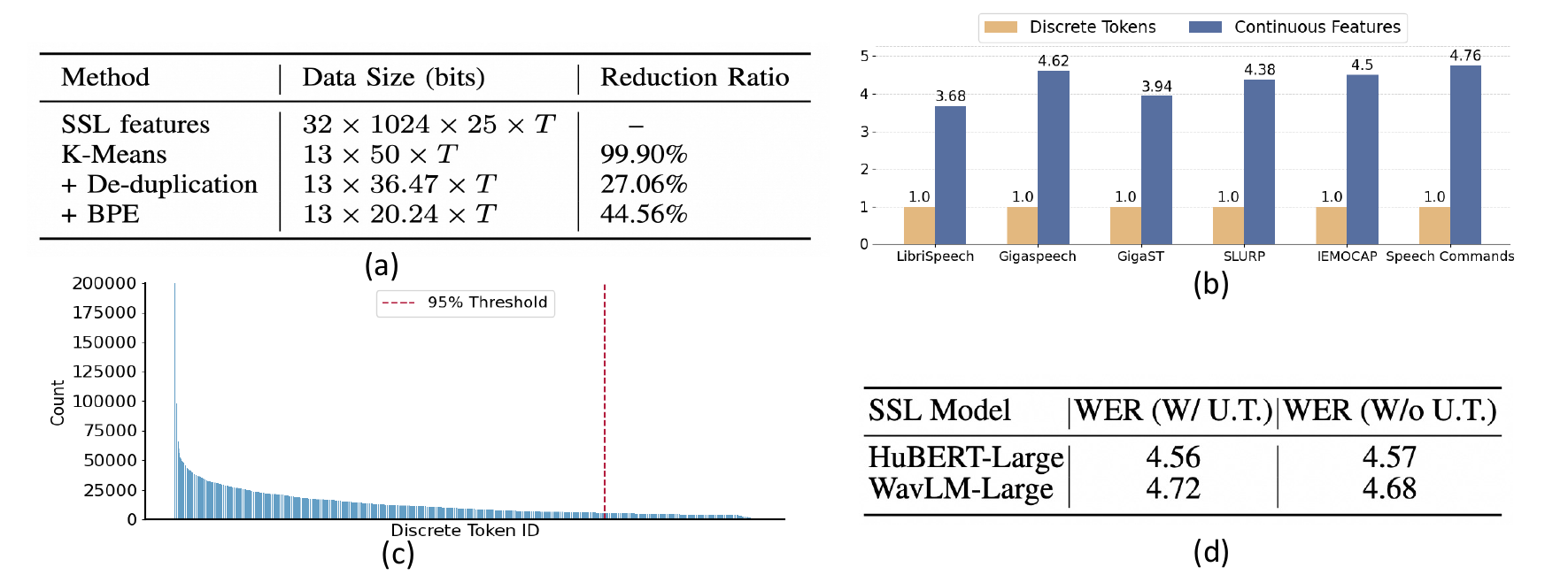}
    \caption{\small Efficiency analysis. (a) Data size (bit) comparison of a T-second utterance among: 1) SSL-based features using HuBERT-Large; 2) discrete tokens in 13-bit with 25 frames/sec. Reduction Ratio is calculated compared to the previous row based on LibriSpeech-100h. (b) Total training time until convergence for discrete tokens and continuous features, with discrete token training time normalized to 1 for all datasets. (c) Frequency distribution of discrete tokens with a codebook size of 6000, based on the GigaSpeech M-size corpus. The red line indicates the 95\% cumulative frequency threshold. (d) WER comparison with and without under-trained tokens (U.T.)}
    \label{fig:efficiency}
\end{figure*}

For SpeechLLMs, the practical cost of a representation is its \emph{bit-rate} \(R=\log_{2}V \cdot C \cdot R_{s}\),  where \(V\) is vocabulary size, \(C\) the number of codebooks, and \(R_{s}\) the emission rate (codes s\(^{-1}\)). Fig.~\ref{fig:efficiency}(a) compares the data size required to represent a \( T \)-second speech utterance using continuous SSL features and compressed discrete tokens. Continuous features from HuBERT-Large (same as WavLM-Large) require \( 32 \times 1024 \times 25 \times T \) bits, where 32 is the bit depth, 1024 is the feature dimensionality, and 25 is the frame rate. In contrast, after K-means compression, this is reduced to \( 13 \times 50 \times T \) bits. Further reductions can be achieved through de-duplication and BPE subword modelling, as calculated based on the LibriSpeech 100-hour dataset. 

This analysis shows that continuous SSL features require orders-of-magnitude more bits than their discrete counterparts. However, bit-rate is only a proxy for Shannon information; 32-bit embeddings exhibit substantial numerical redundancy. The 99.9\% reduction in bit-rate does not strictly correspond to the same level of information loss, but instead indicates an efficient compression of information for discrete tokens. As shown across all tasks, the discrete pipeline incurs only a modest accuracy drop, far smaller than the three-order-of-magnitude gain in bandwidth efficiency (see Table~\ref{tab:main-qwen}). Hence, discrete tokens offer an exceptionally bandwidth-efficient representation, making them attractive for on-device inference, low-bit-rate transmission, and large-scale pre-training where storage or I/O bandwidth is the principal bottleneck.

\paragraph{Training Efficiency}
We maintain consistent training settings (e.g. learning rate schedule) to ensure a fair comparison. Fig.~\ref{fig:efficiency}(b) shows the total training time for convergence with discrete tokens and continuous features across datasets, and the training time for discrete tokens is normalized to 1 for comparison. Our experiments reveal that discrete tokens converge in 4 to 5 epochs, while continuous features take 10 to 15 epochs. In each epoch, discrete tokens also train faster, as their shorter input sequences reduce the computational load, resulting in faster forward/backward passes and lighter token-embedding lookups. This results in a significant reduction in total training time, requiring only 21\% to 27\% of the time needed for continuous features. Discrete tokens’ compact representation not only accelerates the training process but also reduces computational resource demands, making them a more training-efficient choice.

\paragraph{Utility Efficiency}

We conduct utility efficiency analysis for discrete tokens as shown in Fig.~\ref{fig:efficiency}(c), which demonstrates the frequency distribution of 6\,000-size discrete codebook on the GigaSpeech (M size) corpus. Roughly the least-frequent 20\% of tokens together account for only 5\% of all occurrences (this long-tailed pattern we also observe on other datasets). In this context, we follow prior work~\citep{land-bartolo-2024-fishing} in labelling these infrequent tokens as under-trained tokens. These tokens exist in the LLM’s vocabulary and are included during the training process, but are not sufficiently seen. 

This imbalance exposes a key challenge of discrete tokenization: a large portion of the codebook is under-utilized, leading to under-trained tokens that the model rarely, if ever, encounters. Because the LLM codebook reserves fixed vocabulary slots for these under-utilized tokens, capacity that could support higher-utility tokens is wasted. The skewed distribution further injects noise into learning: the model is optimised mainly on the head of the distribution and receives little gradient signal for the tail, which may degrades generalisation—especially for speech segments whose acoustics are mapped to poorly trained codes. 

As shown in Fig.~\ref{fig:efficiency}(d), when we remove 10\% of under-trained tokens during the LibriSpeech 960-hour training process, the WER on the LibriSpeech test-clean set remains roughly unchanged. This suggests that under-trained tokens have a limited effect on performance. These tokens occupy space in the codebook with little gain in performance. By contrast, continuous embeddings provide dense, frame-level representations that capture fine-grained acoustic and linguistic cues without suffering from sparse token utilization.

\begin{figure*}[htbp]
    \centering
    % First subfigure
    \begin{minipage}{0.48\textwidth}
        \centering
        \includegraphics[width=\textwidth]{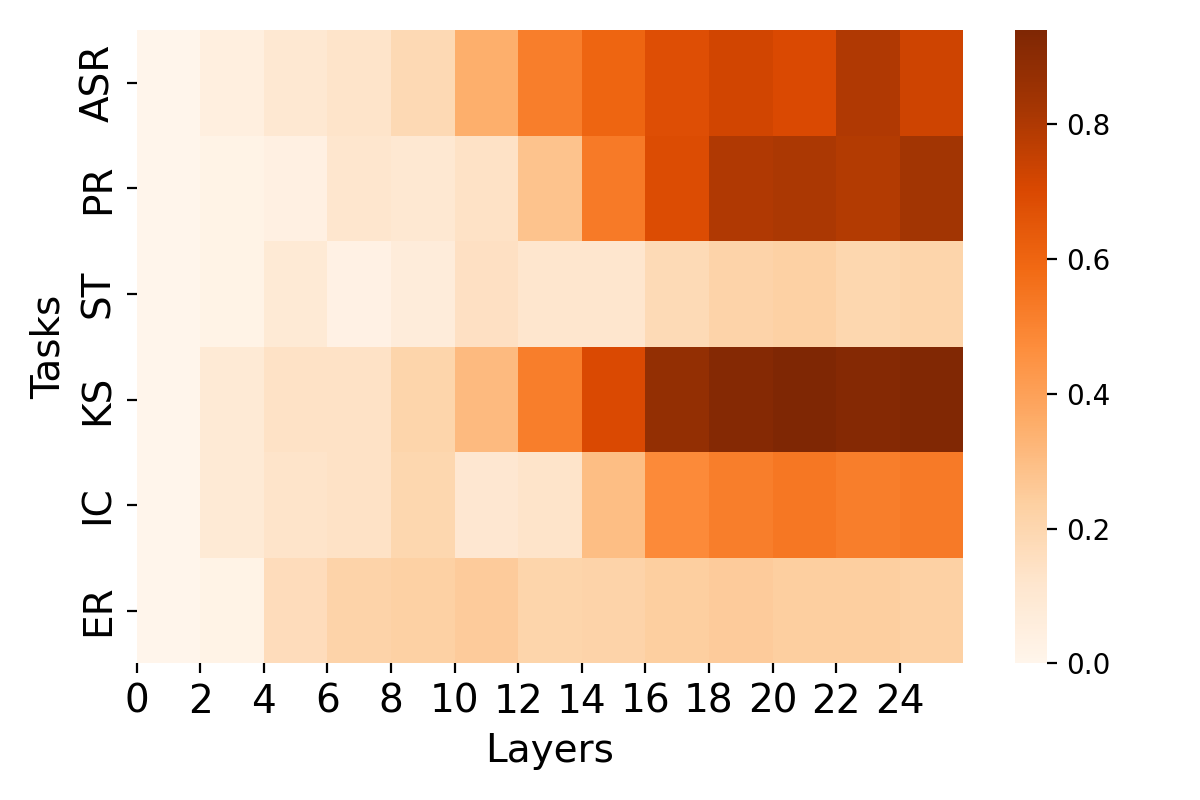}  % Replace with your file
        % \subcaption{Perception} \label{fig:subfig1}
    \end{minipage} \hfill % Adjust horizontal space
    % Second subfigure
    \begin{minipage}{0.48\textwidth}
        \centering
        \includegraphics[width=\textwidth]{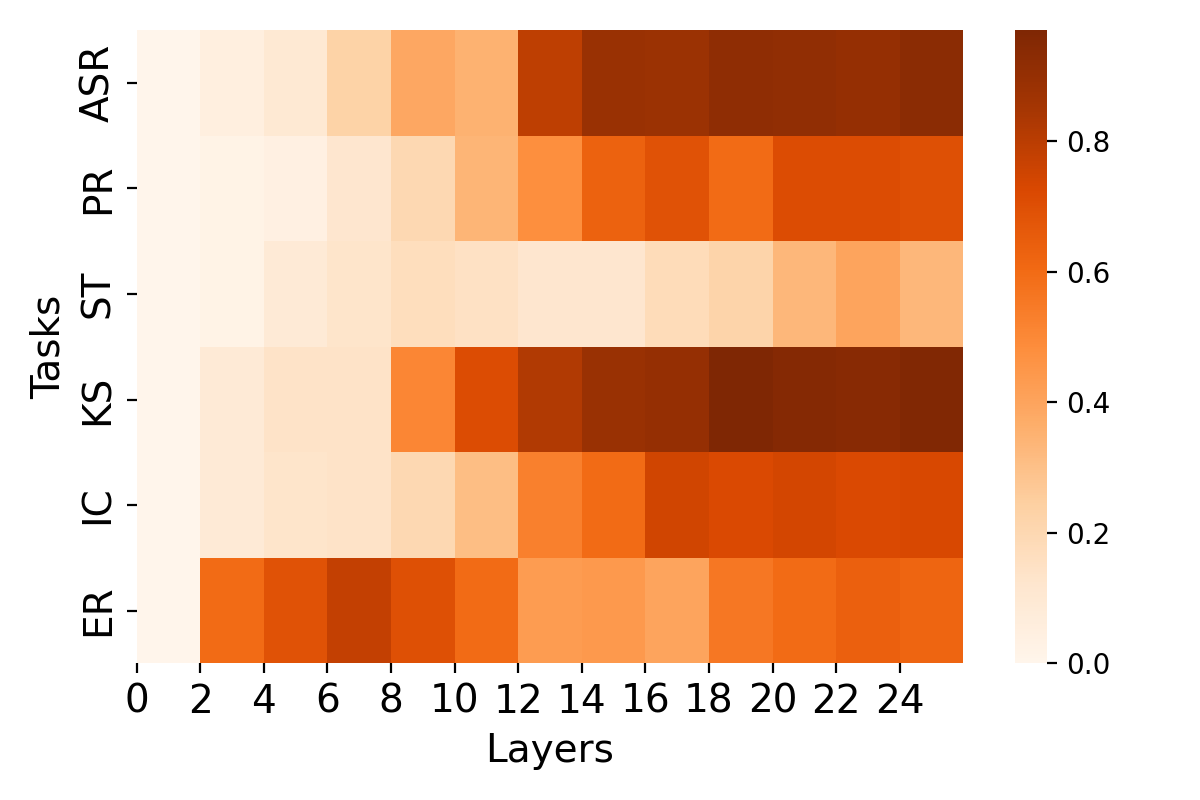}  % Replace with your file
        % #\subcaption{Reasoning} \label{fig:subfig2}
    \end{minipage} \hfill % Adjust horizontal space
    
    % Main caption for the whole figure
    \caption{\small Layer analysis for each task on HuBERT-Large. (Left) Distribution of discrete token performance across each layer. (Right) Distribution of continuous features' performance across each layer.}
    \label{fig:ssl_layer}
    \vspace{-4mm}
\end{figure*}

\subsection{SSL Layer Analysis}

We analyze the performance of discrete tokens and continuous speech embeddings from HuBERT-Large across various tasks. From the results in Fig.~\ref{fig:ssl_layer}, we observe notable trends across different layers. For both discrete tokens and continuous features, we observe similar patterns in phonetic tasks such as ASR and phoneme recognition, where performance is strongest in the deeper layers. A similar trend is observed in semantic tasks like intent classification, suggesting that the model relies increasingly on deeper layers to capture semantic and content-related information. In contrast, emotion recognition exhibits a distinct pattern: continuous features achieve stronger performance in the shallower layers, particularly around layers 4 to 6. It indicates that the model can learn paralinguistic emotion information with the bottom layers. Notably, discrete tokens from K-means quantization fail to perform well in this task, indicating that this method is limited in capturing fine-grained emotional information. These findings offer preliminary insights into the functional roles of SSL layers, guiding future work on interpretability and task-specific SpeechLLMs.

\subsection{LLM Layer Analysis}

\begin{figure}[!t]
    \centering
    \includegraphics[width=1\linewidth]{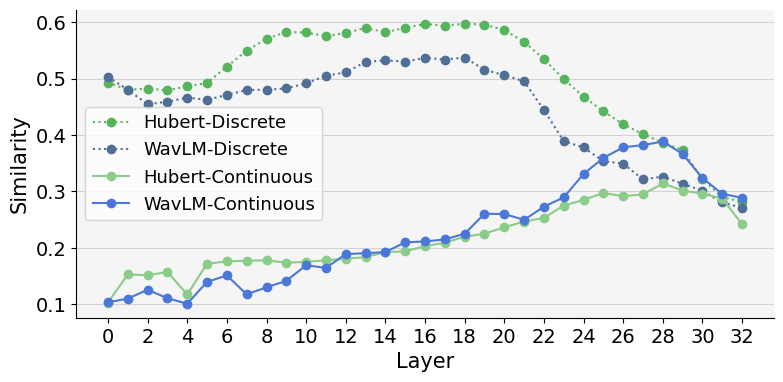}
    \caption{\small Alignments of features obtained from text-form and speech-form inputs.}
    \label{fig:llm_layer}
    \vspace{-2mm}
\end{figure}

Fig.~\ref{fig:llm_layer} illustrates the alignment of features extracted from speech and text inputs over the same words across different decoder layers of the Llama 3.1-8B model. The figure compares the maximum pairwise cosine similarity between speech input (discrete token or continuous feature representations) and text-form sequences across different LLM layers. Interestingly, the trends for discrete tokens and continuous features differ significantly.

For discrete tokens, the similarity between speech and text sequences increases up to layer 22 before decreasing. This suggests that the model initially learns to map discrete speech with text in a manner more similar to text-based representations, but as the layers progress, this alignment starts to diminish. In contrast, for continuous features, the similarity continually rises until around layer 28, where it reaches its peak. This pattern may point to a difference in how the LLM model handles discrete versus continuous representations. With discrete tokens, speech and text seem to be processed as relatively similar modalities: token-level cues highlighting textual similarity at earlier layers. In contrast, continuous features appear to support a more gradual layer-by-layer alignment, suggesting a smoother transition between spoken and written forms throughout the network.

\subsection{Robustness Comparison}

\input{tables/strategy}

We evaluate the robustness of discrete tokens and continuous features based on HuBERT-Large, within the Qwen-1.5-0.5B model on two noisy datasets. SLURP combines varied accents (e.g., Indian English) with a mix of close and distant microphone recordings, while CHiME-4 introduces diverse background environments. The complex acoustic contamination in these datasets challenges the model's noise robustness, especially for discrete tokens. According to our main results, discrete features consistently perform significantly worse than continuous features on both datasets (Table~\ref{tab:main-qwen}).

We hypothesize that the suboptimal performance of discrete tokens in intent classification on the SLURP dataset is due to their limited robustness in basic phonetic perception. To address this, as shown in Table~\ref{tab:strategy}, we first incorporate SLURP transcriptions into ASR training, followed by fine-tuning on intent classification. This approach leads to a substantial improvement in intent classification performance for discrete tokens (accuracy increases from 57.04\% to 66.36\%), with additional gains observed when incorporating the LibriSpeech 100-hour dataset. Similarly, in the CHiME-4 dataset, the zero-shot performance for discrete tokens is suboptimal when evaluated on a model trained solely with LibriSpeech 960-hour ASR data. The performance improved significantly as we progressively incorporated CHiME-4 data along with additional ASR training data. In contrast, for continuous features, the performance improvements from additional training data augmentation are minimal, which indicates less sensitivity to training conditions and a more stable performance across varying data inputs.

%% file: tables/strategy.tex
\begin{table}[!t]
\centering
\footnotesize
\vspace{-0.2cm}
\renewcommand{\arraystretch}{0.8}  % 更小的值，紧凑行间距
\setlength{\tabcolsep}{2pt}  % Adjust column spacing
\begin{tabular}{lc|>{\centering\arraybackslash}p{2cm}}  % Use p{} to adjust only the ACC column width and center contents
\toprule
\multirow{2}{*}{\textit{Datasets}} & \multirow{2}{*}{\textit{Training strategy}} & \multicolumn{1}{c}{Performance} \\ 
\cmidrule{3-3}
& & test \\  % Ensure "test" is centered
\midrule

\multicolumn{2}{c}{\textit{Discrete Tokens}} \\
\midrule

SLURP & Baseline & 57.04 \\
& \hspace{0.5cm}+SLURP ASR & 66.36 \\
& \hspace{1cm}+Libri-100 & 67.18 \\
\midrule

CHiME4 & Zero Shot & 20.35 \\ & Baseline & 16.76 \\
& \hspace{0.5cm}+Libri-960 & 11.37 \\
\midrule

\multicolumn{2}{c}{\textit{Continuous Features}} \\
\midrule

SLURP & Baseline & 76.84 \\
& \hspace{0.5cm}+SLURP ASR & 81.15 \\
& \hspace{1cm}+Libri-100 & 80.86 \\
\midrule

CHiME4 & Zero Shot & 10.81 \\ & Baseline & 8.62 \\
& \hspace{0.5cm}+Libri-960 & 6.59 \\

\bottomrule
\end{tabular}
\caption{\small Robustness analysis with different training strategies: The baseline for SLURP utilizes SLURP data for Intent Classification (IC), while the baseline for CHiME4 employs CHiME4 data for ASR training. "+SLURP ASR" incorporates ASR pretraining on SLURP transcripts, and "+Libri-100" adds 100 hours of LibriSpeech data. For IC, performance is measured by accuracy, and for ASR, by word error rate (WER).}
\label{tab:strategy}
\end{table}

%% file: sections/6-conclusion.tex
\section{Relate Works}

Following the terminology from prior works~\citep{zhang2023speechtokenizer,borsos2023audiolm}, discrete tokens are categorized into two main types: Acoustic tokens and Semantic tokens. Acoustic tokens (e.g., WavTokenizer~\citep{ji2025wavtokenizer}, Soundstream~\citep{zeghidour2021soundstream}, Encodec~\citep{defossez2022high} and ALMTokenizer~\citep{yang2025almtokenizer}) are obtained using compression-based methods, which rely on encoder-decoder architectures with residual vector quantization (RVQ)~\citep{zeghidour2021soundstream}. Semantic tokens~\citep{hsu2021hubert,chen2022wavlm} use clustering algorithms like K-means on SSL model features, with cluster indices serving as discrete representations. In the literature, both semantic and acoustic tokenizers are widely employed in SpeechLLMs. We adopt semantic tokens in our experiments because they enable a more direct and fair comparison with continuous features. Unlike acoustic tokens, which rely on complex compression methods, semantic tokens are derived directly from SSL models—the same source used for continuous features—allowing for a controlled comparison without additional confounding factors.

Most of the prior comparisons between discrete tokens and continuous features have been conducted within traditional speech processing settings, such as ASR and TTS. Works like ~\citep{van_Niekerk_2022, mousavi2024should, chang2024exploring, sharma2022unifyingdiscretecontinuousemotion} are grounded in conventional architectures rather than the multimodal framework of SpeechLLMs, making their findings less directly applicable to this context. Existing comparison analysis research on SpeechLLMs typically focuses on task-specific evaluations~\citep{li2025continuousspeechtokenizertext,xu2024comparingdiscretecontinuousspace}, especially in ASR~\citep{xu2024comparingdiscretecontinuousspace}, without benchmarking across a broader range of spoken language understanding tasks, which is an essential step toward enabling effective human-machine interaction. Moreover, most related studies only focus on surface-level comparisons, lacking deeper analysis of internal mechanisms or cross-task implications. To address this gap, our study evaluates six SLU-related tasks across seven datasets using two SSL models and LLMs of different scales, providing a comprehensive analysis of the characteristics of discrete tokens and continuous features within the SpeechLLMs framework.

\label{4-conclusion}
\section{Conclusion}

This work provides a comprehensive comparison benchmark of SSL-based discrete and continuous speech features across six spoken language understanding tasks: ASR, ST, KS, IC, ER, and PR. Our findings show that, regardless of whether using HuBERT-Large or WavLM-Large, continuous features generally outperform discrete tokens across various LLM backbone scales, except in phoneme recognition. Additionally, our experiments reveal distinct processing patterns for discrete tokens and continuous features in both SSL and LLM layers. Discrete tokens offer advantages in data and training efficiency but show potential for improvement in robustness and utility efficiency. We hope our findings provide valuable insights for the future development of SpeechLLMs.

%% file: sections/7-appendix.tex
\section{Appendix}
\label{sec:appendix}

\subsection{Task Definitions and Prompt Design}

In our instruction-tuning experiments, we select six tasks that span a wide range of spoken language understanding capabilities. Each task is associated with a specific prompt that guides the model to produce the desired output. The details of each task and its corresponding prompt are as follows:

\begin{itemize}
    \item \textbf{Automatic Speech Recognition (ASR)}: The task of converting spoken language into written text. ASR systems are essential for transforming speech into usable data for various applications.  
    \textit{Prompt}: \textit{"Identify the text corresponding to the speech."}
    
    \item \textbf{Phoneme Recognition (PR)}: Involves identifying and classifying the smallest sound units, or phonemes, in speech. This task is fundamental for speech understanding, especially for speech-to-text systems.  
    \textit{Prompt}: \textit{"Identify the phonemes corresponding to the speech."}
    
    \item \textbf{Keyword Spotting (KS)}: The goal is to detect specific predefined words or phrases within an audio stream. This task is important for applications like voice assistants, where the system must recognize specific commands.  
    \textit{Prompt}: \textit{"Identify the keyword corresponding to the speech."}
    
    \item \textbf{Emotion Recognition (ER)}: Focuses on identifying and classifying the emotional state of a speaker based on their speech patterns. This task is crucial for understanding the speaker's emotional tone, often applied in customer service or mental health assessments.  
    \textit{Prompt}: \textit{"Identify the emotion corresponding to the speech."}
    
    \item \textbf{Spoken Intent Classification (IC)}: The objective is to determine the speaker’s intended action or meaning from their speech. This task is commonly used in conversational agents to understand user commands or queries.  
    \textit{Prompt}: \textit{"Identify the intention corresponding to the speech."}
    
    \item \textbf{Speech Translation (ST)}: Converts spoken language from one language into another. This task is essential for real-time translation services, such as simultaneous interpretation or voice translation applications.  
    \textit{Prompt}: \textit{"Translate the English speech into Chinese (or German) text."}
\end{itemize}

Each task in our study is approached with tailored prompts, designed to help the model understand the task's specific requirements and produce accurate results. The prompt design ensures that each task is appropriately framed for instruction tuning, which plays a key role in adapting the model to each task's unique needs.

\subsection{Datasets Used for Instruction Tuning}

In our experiments, we utilize the following datasets for training and evaluating each task:

\textbf{LibriSpeech}~\citep{panayotov2015librispeech}: A large-scale corpus of approximately 1000 hours of 16kHz sampled English read speech, sourced from the LibriVox audiobook project. The dataset is carefully segmented and aligned, widely used for Automatic Speech Recognition (ASR) tasks.

\textbf{GigaSpeech}~\citep{chen2021gigaspeech}: A large-scale, multi-domain English speech recognition dataset containing approximately 10,000 hours of labelled audio. It covers a variety of accents and environments, making it suitable for training diverse speech recognition models.

\textbf{CHiME-4}~\citep{vincent2016chime}: The CHiME-4 dataset is designed for evaluating automatic speech recognition (ASR) in noisy environments. It includes both real and simulated noisy speech data, recorded in four challenging acoustic settings: bus, café, pedestrian area, and street junction. The dataset features speech from the Wall Street Journal (WSJ0) corpus, captured using a 6-channel microphone array in real-world noise conditions.

\textbf{IEMOCAP}~\citep{busso2008iemocap}: A dataset containing 151 dialogue sessions between actors, annotated with 9 emotion labels (e.g., anger, excitement, fear, sadness). This dataset is used for emotion recognition tasks, providing rich samples of emotional speech.

\textbf{SLURP}~\citep{bastianelli2020slurp}: A task-oriented spoken language understanding dataset containing dialogue data from 18 different domains, primarily used for intent classification tasks. Each speech sample is labelled with an intent category, which helps the model classify user intent from spoken commands.

\textbf{GigaST}~\citep{chen2021gigaspeech}: A large-scale speech translation corpus created by translating the transcribed text of GigaSpeech into German and Chinese. The training data is machine-translated, while the test set is manually translated, making it suitable for speech translation tasks.

\textbf{Speech Commands}~\citep{warden2018speech}: provided by Google, consists of 35 short spoken keywords with over 100,000 audio samples collected from 2,600 speakers. It is primarily used for keyword spotting tasks and includes labeled 1-second .wav files recorded in English.

\subsection{One Epoch Training Efficiency Comparison}

\begin{figure}[!t]
    \centering
    \includegraphics[width=0.45\textwidth]{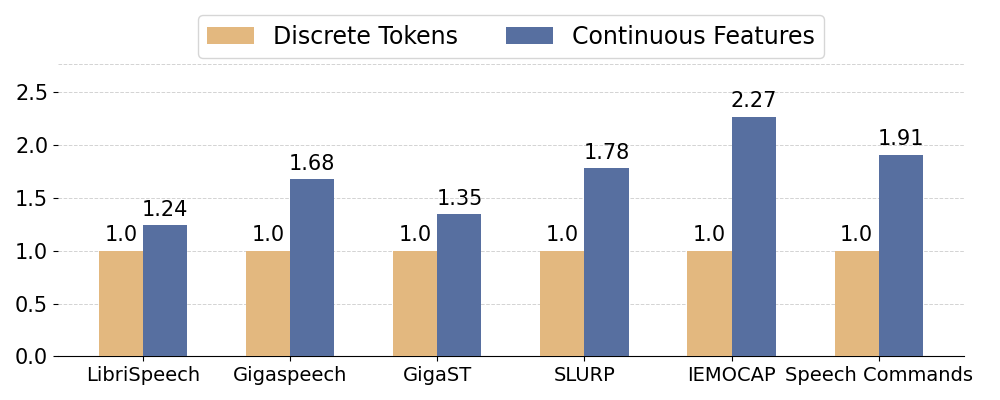}
    \caption{Training time per epoch using SSL continuous / discrete tokens. We normalize the training time of discrete tokens to unit 1 for all datasets}
    \label{fig:training_time_epoch}
\end{figure}

Fig.~\ref{fig:training_time_epoch} highlights the training time for one epoch across the datasets. After normalizing the training time of discrete tokens to 1, we observe that the time required for training with discrete tokens is between 33\% and 67\% less than that of continuous features. This further underscores the efficiency of discrete tokens, especially in terms of both time and resource usage.

